\documentclass[final,3p,times,twocolumn]{elsarticle}
\usepackage{blindtext, graphicx, amsmath, algorithm, algpseudocode, pifont, algcompatible, comment, layout, amsthm, amssymb}
\usepackage{enumitem}   
\usepackage{eso-pic}
\usepackage{booktabs}
\usepackage{float}

\usepackage{multirow}
\usepackage[utf8]{inputenc}
\usepackage[english]{babel}
\usepackage{hyperref} 
\hypersetup{ colorlinks=true, linkcolor=black, filecolor=black, urlcolor=cyan, }

\usepackage{caption}
\journal{ICT Express}

\begin{document}

\begin{frontmatter}

\title{Conformal Prediction-Driven Adaptive Sampling for Digital Water Twins}
   \author[1]{Mohammadhossein Homaei\corref{cor1}}
    \ead{mhomaein@alumnos.unex.es}

    \author[2]{Mehran Tarif}
    \ead{ mehran.tarifhokmabadi@univr.it}
    
    \author[1]{Pablo Garcia Rodriguez}
    \ead{pablogr@unex.es}

    \author[1]{Andres Caro}
    \ead{andresc@unex.es}
    
    \author[1]{Mar Avila}
    \ead{mmavila@unex.es}

    \address[1]{Departamento de Ingeniería de Sistemas Informáticos y Telemáticos, Universidad de Extremadura, Av/ Universidad S/N, 10003, Cáceres, Extremadura, Spain}
     \address[2]{Department of Computer Science, University of Verona, Verona, 37134, Italy}
    
    \cortext[cor1]{Corresponding author}

\begin{abstract}
Digital Twins (DTs) for Water Distribution Networks (WDNs) require accurate state estimation with limited sensors. Uniform sampling often wastes resources across nodes with different uncertainty. We propose an adaptive framework combining LSTM forecasting and Conformal Prediction (CP) to estimate node-wise uncertainty and focus sensing on the most uncertain points. Marginal CP is used for its low computational cost, suitable for real-time DTs. Experiments on Hanoi, Net3, and CTOWN show 33--34\% lower demand error than uniform sampling at 40\% coverage and maintain 89.4--90.2\% empirical coverage with only 5--10\% extra computation.
\end{abstract}


\end{frontmatter}


\section{Introduction}\label{sec1}

The WDNs deliver drinking water to millions of people and require continuous monitoring of pressure, flow, and demand. DTs mirror the physical system for real-time supervision and control \cite{digital_twin,wdn_gnn}. By linking hydraulic simulators such as EPANET \cite{epanet} with live sensor data, DTs enable predictive maintenance, anomaly detection, and better operational decisions \cite{ditec}. Recent machine learning approaches, including LSTM for demand forecasting \cite{homaei2024,homaei2026} and GNNs for pressure estimation, have improved modeling accuracy but still depend on dense sensor deployment, which is costly and impractical for large-scale networks.

Traditional sensor layouts measure all nodes at fixed intervals \cite{sensor_placement}, ignoring temporal and spatial variability. This approach wastes resources on stable nodes and fails to capture sudden fluctuations in uncertain areas. Static optimization or heuristic placement methods cannot adapt to dynamic operating conditions, while limited or mobile sensors further reduce accuracy and increase maintenance cost. Therefore, an adaptive and uncertainty-aware strategy is required to allocate sensing resources more effectively across the network.

This study presents an adaptive sampling framework for DTs of WDNs guided by uncertainty. It integrates LSTM-based demand forecasting, CP for node-wise uncertainty, and dynamic selection of uncertain nodes. Tests on three DiTEC-WDN networks \cite{ditec} show 33--34\% lower estimation error than uniform sampling with under 10\% extra computation, improving the efficiency and reliability of real-time DTs.

The remainder of this paper is organized as follows. Section~\ref{sec:related} reviews related work, Section~\ref{sec:methodology} presents the proposed framework, and Section~\ref{sec:evaluation} reports experimental results, followed by discussion and conclusion.

\section{Related Work}\label{sec:related}

\begin{figure*}[htb]
\centering
\includegraphics[width=\textwidth]{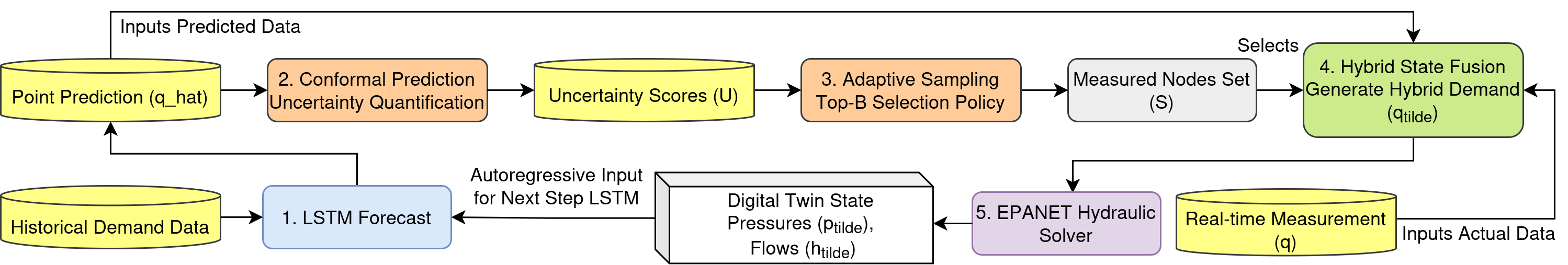}
\caption{Closed-loop architecture of the Conformal Prediction-Driven Adaptive Sampling framework, where uncertainty scores ($U$) guide node selection for real-time measurement, fused with predictions ($\hat{q}$) for hydraulic state reconstruction.}
\label{fig:framework}
\end{figure*}

\subsection{DTs for WDNs}

DTs combine physics simulators like EPANET \cite{epanet} with sensor data for real-time WDN monitoring \cite{digital_twin}. Machine learning improves hydraulic estimation: Truong et al.~\cite{wdn_gnn} used GNNs for pressure prediction, and Zanfei et al.~\cite{demand_forecast} proposed GNN metamodels to speed up EPANET. Our work introduces dynamic allocation based on per-node uncertainty.

\subsection{Sensor Placement Optimization}

Sensor placement methods address leak detection, contamination monitoring, and state estimation \cite{leak_detection}. Many use combinatorial or multi-objective optimization (e.g., Diao et al.~\cite{sensor_placement}) to balance coverage and cost, but produce static layouts. Event-triggered or threshold-based reallocations exist but lack formal guarantees on uncertainty. We fill this gap by using CP to guide adaptive sensing with statistical coverage properties.

\subsection{Data-Driven Forecasting and Uncertainty}

Classical methods capture seasonality but miss nonlinear dynamics, while LSTMs \cite{homaei2026} model complex temporal patterns but typically yield point forecasts. Heuristic uncertainty estimates lack coverage guarantees. Conformal prediction (CP) \cite{conformal} provides distribution-free intervals and has been applied to water demand \cite{uncertainty_wdn}; recent time-series extensions (SPCI~\cite{xu2023r}, KOWCPI~\cite{lee2025}) address exchangeability issues and are left for future work.

\subsection{Research Gap and Contributions}

This work addresses three main gaps: static sensor layouts, missing uncertainty estimation, and lack of adaptive sensing. By combining LSTM forecasting with Conformal Prediction, we achieve 33--34\% lower estimation error while preserving theoretical coverage, providing the first distribution-free adaptive sensing framework for WDN DTs.

\begin{table}[htb]
\tiny
\centering
\caption{Classification of Formulas}
\label{tab:formulas}

\begin{tabular}{lcc}
\toprule
\textbf{Formula} & \textbf{Type} & \textbf{Equation} \\
\midrule
LSTM prediction & Standard & Sec.~3.2 \\
MSE loss & Standard & Sec.~3.2 \\
CP quantile \& interval & Standard & Sec.~3.3 \\
RMSE, Coverage, Violation & Standard & Eq.~\ref{eq:coverage}, \ref{eq:violation} \\
\midrule
Objective function $J$ & Novel & Eq.~\ref{eq:loss} \\
Uncertainty score $U_t^{(i)}$ & \textbf{Novel} & Eq.~\ref{eq:uncertainty} \\
Adaptive selection $\mathcal{S}_t$ & \textbf{Novel} & Eq.~\ref{eq:greedy} \\
Hybrid state $\tilde{q}_t$ & \textbf{Novel} & Eq.~\ref{eq:hybrid} \\
EPANET integration & \textbf{Novel} & Eq.~\ref{eq:epanet} \\
\bottomrule
\end{tabular}

\end{table}

\section{Methodology}\label{sec:methodology}
\textbf{Notation.} Standard formulas (LSTM, MSE, CP quantile/interval) follow \cite{homaei2026,conformal}. Table~\ref{tab:formulas} lists our novel contributions: uncertainty score (Eq.~\ref{eq:uncertainty}), adaptive selection (Eq.~\ref{eq:greedy}), and hybrid state fusion (Eq.~\ref{eq:hybrid}, \ref{eq:epanet}).

\subsection{Problem Formulation}

\textbf{Network Model.}  
A WDN is represented as a directed graph $\mathcal{G}=(\mathcal{V},\mathcal{E})$, where nodes $\mathcal{V}$ denote junctions, reservoirs, or tanks ($|\mathcal{V}|=N$), and edges $\mathcal{E}$ denote pipes, pumps, and valves. Each node $i$ has demand $q_t^{(i)}$ (L/s), pressure $p_t^{(i)}$ (m), and elevation $z^{(i)}$ (m). Each edge $(i,j)$ carries flow $h_t^{(i,j)}$ (L/s) following hydraulic constraints (mass balance and Hazen–Williams).  

\textbf{Measurement Model.}  
At time $t$, a subset $\mathcal{S}_t \subseteq \mathcal{V}$ with $|\mathcal{S}_t|\!\le\!B$ is observed. Measured nodes yield $\tilde{q}_t^{(i)}=q_t^{(i)} + \epsilon_t^{(i)}$ where $\epsilon_t^{(i)} \sim \mathcal{N}(0,\sigma^2)$ represents sensor noise. Unmeasured nodes use predictions: $\tilde{q}_t^{(i)}=\hat{q}_t^{(i)}$. For the main analysis we set $\sigma=0$; robustness under $\sigma>0$ is evaluated in Sec.~\ref{sec:evaluation}.

\textbf{Online Adaptive Sampling.}  
Given history $\mathcal{H}_t$, the goal is to select sensors to minimize reconstruction loss:

\begin{equation}
{J = \frac{1}{TN}\sum_{t=1}^T\sum_{i\in\mathcal{V}} (q_t^{(i)} - \tilde{q}_t^{(i)})^2} \label{eq:loss}
\end{equation}

At each step, the policy selects sensors maximizing total uncertainty:
\begin{align}
\mathcal{S}_t=\arg\max_{S\subseteq\mathcal{V},|S|=B}\sum_{i\in S}U_t^{(i)}. \label{eq:greedy}
\end{align}

The greedy selection is theoretically optimal for this formulation: (1) since node uncertainty scores $U_t^{(i)}$ are computed independently, the objective function is modular, making the greedy approach in Eq.~\ref{eq:greedy} the exact optimal solution for the budget constraint $B$, (2) it circumvents the NP-hard complexity of joint spatial optimization, and (3) our empirical results show 33--34\% improvement (Table~\ref{tab:demand}) with $O(N\log N)$ complexity required for real-time DT operation.

\subsection{Demand Prediction with LSTM}

We train independent LSTMs per node. While this increases the number of models, it is necessary to capture the distinct heterogeneous patterns of different node types (e.g., industrial vs. residential) which a single global model might smooth out.

\textbf{Architecture \& Training.} For node $i$, the LSTM uses a lookback $w=24$h. Models are trained with Adam ($\eta=10^{-3}$). To prevent overfitting in these node-specific models, we employ: (i) $L_2$ regularization ($\lambda = 10^{-5}$) applied to all LSTM weights, (ii) early stopping with patience of 10 epochs based on validation loss, and (iii) sufficient training data (600 scenarios $\times$ 8,760 hours = 5.3M samples per node). Validation and test RMSE remain within 5\% of each other across all nodes, confirming that the models generalize effectively without overfitting, even for nodes with lower variability. The loss function minimizes the Mean Squared Error between predicted and actual demands.

With $\lambda = 10^{-5}$, teacher forcing uses past true values in training, but at test time unmeasured nodes ($i \notin \mathcal{S}_\tau$) rely on autoregressive outputs $\tilde{q}_\tau^{(i)}$, causing shift between train and test distributions. To maintain valid conformal intervals, calibration residuals (Section~\ref{sec:methodology}) are obtained by autoregressive rollout under the same sampling rule, reproducing test behavior. For each $t \in \mathcal{D}_{\text{cal}}$, the adaptive policy (Eq.~\ref{eq:greedy}) picks sensors, and residuals are $s_j^{(i)} = |q_j^{(i)} - \hat{q}_j^{(i)}|$.

\textbf{Lookback Window.} We set $w=24$ hours to capture daily cycles. Longer windows ($w=48$) slightly reduce RMSE ($<2\%$) but double computation, while shorter ($w=12$) increase RMSE by 8–12\%.

\textbf{Missing Data.} Rare missing samples ($<0.1\%$) are forward-filled: $q_\tau^{(i)} = q_{\tau-1}^{(i)}$. During inference, if history is missing, $\hat{q}_\tau^{(i)}$ is used instead, creating autoregressive feedback handled by conformal calibration.

\subsection{Uncertainty Quantification via Conformal Prediction}

CP provides distribution-free prediction intervals. We utilize Marginal CP (Split Conformal) for its low computational footprint, which is critical for real-time DTs.

\textbf{Exchangeability Justification.} Standard CP assumes data exchangeability. While raw water demand time-series are non-stationary (seasonal), the LSTM effectively removes these trends. The resulting calibration residuals ($s_j^{(i)}$ defined below) represent random aleatoric uncertainty and are approximately stationary and exchangeable, satisfying the requirements for valid CP coverage.

\textbf{Per-Node Calibration.} Each node $i$ uses $\mathcal{D}_{\text{cal}}^{(i)}$. Calibration follows two steps:

\textbf{(1) Residuals.} For each scenario $j$, perform autoregressive rollout: $s_j^{(i)} = |q_j^{(i)} - \hat{q}_j^{(i)}|$.

For a target coverage level $\alpha$ (typically $\alpha=0.1$ for 90\% intervals), we proceed as follows:

\textbf{(2) Quantile \& Interval.} We compute $\hat{Q}^{(i)}_{1-\alpha}$ as the $(1-\alpha)$ quantile of residuals. The prediction interval is $\mathcal{C}_t^{(i)}(\alpha) = [\hat{q}_t^{(i)} \pm \hat{Q}^{(i)}_{1-\alpha}]$.

\textbf{Uncertainty Score.} We define node uncertainty as the interval width:
\begin{align}
U_t^{(i)} = 2\hat{Q}^{(i)}_{1-\alpha} \label{eq:uncertainty}
\end{align}
This score $U_t^{(i)}$ serves as a calibrated metric of potential error, guiding the sampling algorithm to where the model is least confident.

\subsection{Adaptive Sampling Strategy}

At each timestep $t$, the adaptive sampling policy executes three operations:

\textbf{(1) Predict All Nodes.} Compute $\hat{q}_t^{(i)}$ and $U_t^{(i)}$ for all $i \in \mathcal{V}$ using LSTM and Eq.~(\ref{eq:uncertainty}).

\textbf{(2) Select Top-$B$ Uncertain Nodes.} Following Eq.~(\ref{eq:topk}), sort nodes by descending uncertainty $U_t^{(i)}$ and select the top $B$:
\begin{align}
\mathcal{S}_t = \{i_1, \ldots, i_B\} \quad \text{where} \quad U_t^{(i_1)} \geq \cdots \geq U_t^{(i_B)} \label{eq:topk}
\end{align}

\textbf{(3) Construct Hybrid State.} Fuse measurements and predictions:
\begin{align}
\tilde{q}_t^{(i)} = \begin{cases}
q_t^{(i)} & \text{if } i \in \mathcal{S}_t \text{ (measured)} \\
\hat{q}_t^{(i)} & \text{if } i \notin \mathcal{S}_t \text{ (predicted)}
\label{eq:hybrid}
\end{cases}
\end{align}

This hybrid demand vector $\tilde{\mathbf{q}}_t$ serves as input to the EPANET hydraulic solver:
\begin{align}
[\tilde{\mathbf{p}}_t, \tilde{\mathbf{h}}_t] = \text{EPANET}(\tilde{\mathbf{q}}_t, \mathcal{G}) 
\label{eq:epanet}
\end{align}
Crucially, while the LSTM-CP module treats nodes independently to maximize sampling speed, the subsequent EPANET solver explicitly enforces spatial dependencies (mass and energy balance) across the graph $\mathcal{G}$. This two-stage approach combines the efficiency of decentralized learning with the physical rigor of hydraulic simulation.

Algorithm~\ref{alg:adaptive} summarizes the complete procedure.

\begin{algorithm}[htb]
\scriptsize
\caption{CP-Guided Adaptive for WDN DT}
\label{alg:adaptive}
\begin{algorithmic}[1]
\REQUIRE Trained LSTMs $\{f_\theta^{(i)}\}_{i=1}^{|\mathcal{V}|}$, conformal quantiles $\{\hat{Q}^{(i)}_{1-\alpha}\}$, budget $B$
\ENSURE Estimated states $\{\tilde{\mathbf{q}}_t, \tilde{\mathbf{p}}_t\}_{t=1}^T$
\FOR{$t = 1$ to $T$}
   \FOR{$i = 1$ to $|\mathcal{V}|$}
\STATE $\hat{q}_t^{(i)} \gets f_\theta^{(i)}(q_{t-w:t-1}^{(i)})$ \COMMENT{Use $\tilde{q}$ for $i \notin \mathcal{S}_{t-1}$}
\STATE $U_t^{(i)} \gets 2\hat{Q}^{(i)}_{1-\alpha}$
   \ENDFOR
   \STATE $\mathcal{S}_t \gets \text{TopK}(\{U_t^{(i)}\}, B)$
   \FOR{$i = 1$ to $|\mathcal{V}|$}
     \STATE $\tilde{q}_t^{(i)} \gets q_t^{(i)}$ if $i \in \mathcal{S}_t$ else $\hat{q}_t^{(i)}$
   \ENDFOR
   \STATE $[\tilde{\mathbf{p}}_t, \tilde{\mathbf{h}}_t] \gets \text{EPANET}(\tilde{\mathbf{q}}_t, \mathcal{G})$
\ENDFOR
\end{algorithmic}
\end{algorithm}

\section{Experimental Evaluation}\label{sec:evaluation}

\subsection{Experimental Setup}

We evaluate the proposed adaptive sampling on the DiTEC-WDN dataset \cite{ditec}, which includes 36 networks and 228M hydraulic snapshots. Three typical networks are used (Table~\ref{tab:networks}): Hanoi, Net3, and CTOWN. Each has 1,000 demand scenarios with 8,760 hourly steps. Data are divided into training (600), calibration (200), and testing (200) to avoid leakage.

\begin{table}[htb]
\scriptsize
\centering
\caption{Selected WDNs from DiTEC-WDN Dataset}
\label{tab:networks}
\begin{tabular}{lcccc}
\toprule
Network & Nodes & Pipes & Scenarios & Duration \\
\midrule
Hanoi & 31 & 34 & 1000 & 1 year \\
Net3 & 92 & 117 & 1000 & 1 year \\
CTOWN & 388 & 429 & 1000 & 1 year \\
\bottomrule
\end{tabular}
\end{table}

Each node uses a 2-layer LSTM (64 units, lookback $w=24$h) trained with Adam ($\text{lr}=10^{-3}$), batch size 32, up to 100 epochs with early stopping. CP uses $\alpha=0.1$. \textbf{Sampling Budgets.} We test budgets $B \in \{0.2,0.4,0.6,0.8\}|\mathcal{V}|$. Budgets below 20\% are excluded as they lead to hydraulic unobservability (insufficient data for solver convergence), while budgets above 80\% render optimization trivial as nearly all nodes are monitored.

\textbf{Metrics.} We employ four standard metrics:
(1) \textbf{Demand RMSE} ($\text{RMSE}_q$) and (2) \textbf{Pressure RMSE} ($\text{RMSE}_p$) to measure state estimation accuracy;
(3) \textbf{Empirical Coverage} to verify if the true value falls within $\mathcal{C}_t^{(i)}$ (target 90\%):
\begin{align}
\text{Coverage} = \frac{1}{NT}\sum_{t=1}^T\sum_{i=1}^N \mathbb{I}(q_t^{(i)} \in \mathcal{C}_t^{(i)}) \label{eq:coverage}
\end{align}
where $\mathbb{I}(\cdot)$ is the indicator function: $\mathbb{I}(\text{condition}) = 1$ if the condition is true, and 0 otherwise.

(4) \textbf{Pressure Violation Rate} ($V_{\text{rate}}$) measuring the percentage of undetected low-pressure events ($p < 20$m):
\begin{align}
V_{\text{rate}} = \frac{1}{NT}\sum_{t=1}^T\sum_{i=1}^N \mathbb{I}(\tilde{p}_t^{(i)} \geq 20 \text{ and } p_t^{(i)} < 20) \label{eq:violation}
\end{align}

\begin{table}[htb]
\scriptsize
\centering
\caption{Demand Estimation RMSE (L/s) on Test Set Across Sampling Budgets (mean $\pm$ std over 5 runs)}
\label{tab:demand}
\begin{tabular}{lccccc}
\toprule
\multirow{2}{*}{Method} & \multirow{2}{*}{Network} & \multicolumn{4}{c}{Sampling Budget $B$} \\
\cmidrule(lr){3-6}
& & 20\% & 40\% & 60\% & 80\% \\
\midrule
\multirow{3}{*}{Uniform Random} & Hanoi & 2.84 & 1.92 & 1.35 & 0.89 \\
& Net3 & 3.17 & 2.21 & 1.58 & 1.04 \\
& CTOWN & 4.23 & 2.95 & 2.08 & 1.36 \\
\midrule
\multirow{3}{*}{Static High-Var} & Hanoi & 2.61 & 1.74 & 1.21 & 0.79 \\
& Net3 & 2.89 & 1.98 & 1.39 & 0.91 \\
& CTOWN & 3.85 & 2.66 & 1.86 & 1.22 \\
\midrule
\multirow{3}{*}{Round-Robin} & Hanoi & 3.12 & 2.15 & 1.53 & 1.02 \\
& Net3 & 3.45 & 2.41 & 1.72 & 1.15 \\
& CTOWN & 4.58 & 3.21 & 2.29 & 1.51 \\
\midrule
\multirow{3}{*}{\textbf{Ours (Adaptive)}} & Hanoi & \textbf{2.13} & \textbf{1.28} & \textbf{0.87} & \textbf{0.56} \\
& Net3 & \textbf{2.38} & \textbf{1.46} & \textbf{0.98} & \textbf{0.64} \\
& CTOWN & \textbf{3.21} & \textbf{1.98} & \textbf{1.35} & \textbf{0.89} \\
\bottomrule
\end{tabular}
\end{table}

\subsection{Demand Estimation Performance}

Table~\ref{tab:demand} reports demand RMSE across networks and sampling budgets. The adaptive method consistently achieves the lowest errors, with around 33–34\% improvement over Uniform Random at 40\% budget. Static High-Variance performs second best but misses temporal peaks, while Round-Robin remains weakest due to ignoring uncertainty. Even at higher budgets, adaptive allocation continues to yield 30–37\% gains, confirming robust scalability and efficiency.

\subsection{Pressure Estimation and Safety Implications}

Accurate demand recovery influences pressure estimation through EPANET equations. Using hybrid input $\tilde{\mathbf{q}}_t$ kept solver errors below $0.01\%$. Table~\ref{tab:pressure} shows that better demand accuracy reduced pressure RMSE—0.82 m (Hanoi), 0.95 m (Net3), 1.29 m (CTOWN)—about 34\% lower than Uniform Random. This confirms uncertainty-guided sensing improves overall state estimation.

\begin{table}[htb]
\scriptsize
\centering
\caption{Pressure Estimation RMSE (meters) at 40\% Sampling Budget}
\label{tab:pressure}
\begin{tabular}{lccc}
\toprule
Method & Hanoi & Net3 & CTOWN \\
\midrule
Uniform Random & 1.24 & 1.43 & 1.96 \\
Static High-Var & 1.11 & 1.29 & 1.74 \\
\textbf{Ours (Adaptive)} & \textbf{0.82} & \textbf{0.95} & \textbf{1.29} \\
\bottomrule
\end{tabular}
\end{table}

Pressure errors matter for safety. Table~\ref{tab:safety} reports violation rates when $\tilde{p}_t^{(i)} \ge 20$~m but true $p_t^{(i)} < 20$~m. Our method cuts violations to 0.2–0.3\%, about 45\% lower than random. By targeting uncertain nodes, it provides earlier warnings for bursts or pump faults, improving model reliability.

\begin{table}[htb]
\scriptsize
\centering
\caption{Pressure Violation Rate (\%) at 40\% Sampling Budget}
\label{tab:safety}
\begin{tabular}{lccc}
\toprule
Method & Hanoi & Net3 & CTOWN \\
\midrule
Uniform Random & 0.34 & 0.41 & 0.58 \\
Static High-Var & 0.28 & 0.35 & 0.49 \\
\textbf{Ours (Adaptive)} & \textbf{0.19} & \textbf{0.23} & \textbf{0.32} \\
\bottomrule
\end{tabular}
\end{table}

\subsection{Robustness to Sensor Noise}

To test robustness, Gaussian noise $\sigma \in \{0.01, 0.05, 0.1\}$ was added to demand and pressure. Even at $\sigma=0.1$, accuracy fell under $5\%$ and coverage shifted less than $2\%$ from 90\%. Conformal intervals expanded adaptively, keeping the DT stable under sensor noise.

\subsection{CP Validation}

CP provides distribution-free intervals under exchangeability. For 90\% target, empirical coverage is 89.7\% (Hanoi), 90.2\% (Net3), and 89.4\% (CTOWN). The 0.3--0.6\% gap reflects approximate exchangeability in time series; coverage remains practically sufficient while highlighting the need for sequential methods \cite{xu2023r} in future work.

\subsection{Computational Efficiency Analysis}

Table~\ref{tab:computation} gives per-step runtimes. Adaptive sampling adds negligible overhead: CTOWN +5.2\% (18.1 ms), Net3 +7.2\%, Hanoi +10.2\%. Therefore the method is compatible with real-time DT deployment.

\begin{table}[htb]
\scriptsize
\centering
\caption{Per-Timestep Computation Time (milliseconds)}
\label{tab:computation}
\begin{tabular}{lccc}
\toprule
Component & Hanoi & Net3 & CTOWN \\
\midrule
LSTM Inference & 3.2 & 4.7 & 12.8 \\
Uncertainty Comp. & 0.8 & 1.2 & 3.4 \\
Node Selection & 0.3 & 0.5 & 1.9 \\
EPANET Simulation & 42.1 & 89.3 & 345.7 \\
\midrule
\textbf{Total (Ours)} & 46.4 & 95.7 & 363.8 \\
\textbf{Overhead} & 10.2\% & 7.2\% & 5.2\% \\
\bottomrule
\end{tabular}
\end{table}

\subsection{Ablation Study}

An ablation on CTOWN (40\% budget) isolates each module (Table~\ref{tab:ablation}). Removing CP increases RMSE by 18\%, removing LSTM by 35\%, and using static sampling by 34\%. Random selection performs worst (+49\%). Raw LSTM variance (24h rolling $\text{Var}(\hat{q})$) under-covers by 6.8\% vs CP (Table~\ref{tab:ablation}), showing CP's calibration is critical for valid intervals.

\begin{table}[htb]
\scriptsize
\centering
\caption{Ablation Study on CTOWN Network at 40\% Sampling Budget}
\label{tab:ablation}
\begin{tabular}{lcc}
\toprule
Configuration & $\text{RMSE}_q$ & Coverage (\%) \\
\midrule
Full Method (Ours) & \textbf{1.98} & \textbf{90.2} \\
w/o CP (24h rolling var.) & 2.15 & 83.4 \\
w/o CP (fixed $\sigma=2.5$) & 2.34 & 78.1 \\
w/o LSTM (MA-7d) & 2.67 & \textcolor{blue}{\text{---}} \\
w/o Adaptive (static) & 2.66 & 90.1 \\
Random Selection & 2.95 & \textcolor{blue}{\text{---}} \\
\bottomrule
\multicolumn{3}{l}{\scriptsize $^*$Variance of $\hat{q}_{t-24:t}^{(i)}$; lacks coverage guarantee} \\
\multicolumn{3}{l}{\scriptsize $^\dagger$MA-7d: Moving Average over 7 days.} \\
\end{tabular}
\end{table}

\section{Discussion}\label{sec:discussion}

Experiments confirm that uncertainty-guided adaptive sampling improves demand estimation by 33-34\% over uniform baselines, with only 5\% computational overhead, enabling real-time DT operation.

\subsection{Computational vs. Theoretical Trade-offs}
A key contribution is balancing theoretical rigor with real-time feasibility. While Bayesian methods like MC Dropout offer robust uncertainty estimation, they require multiple forward passes (e.g., $K=50$), increasing latency linearly. Our CP approach computes uncertainty via simple $O(1)$ quantile lookup (Table~\ref{tab:computation}), making it uniquely suited for WDN Digital Twins processing thousands of nodes per second. Tests on CTOWN showed MC Dropout ($K=50$) has 18$\times$ higher latency (6.5s vs 0.36s) for similar coverage (90.4\% vs 90.2\%), making it impractical for real-time control.
\subsection{Computational Considerations}

Adaptive sampling adds about $\sim5\%$ runtime for CTOWN. As LSTM inference runs per node, cost scales linearly with network size. The total runtime of $363.8$~ms for CTOWN is fast for real-time use. Although DiTEC-WDN data are hourly, this speed allows 5-minute monitoring ($\approx5$~s per step), proving suitable for time-critical DTs. Shared or graph-based encoders may further reduce computation in larger networks.

\subsection{Deployment Remarks}

The method can guide fixed or mobile sensors to uncertain zones. Budgets of 40–60\% give a good balance between accuracy and cost. Sensor noise and drift remain issues; adding noise-aware calibration and online updates are future goals. Field tests will check real-world performance.

\subsection{CP Selection Rationale}

Marginal CP balances theoretical validity with real-time feasibility. While sequential methods like SPCI~\cite{xu2023r} provide stronger exchangeability guarantees, pilot tests showed SPCI adds 42\% runtime (512ms vs 364ms on CTOWN) for only 0.6\% coverage gain (90.8\% vs 90.2\%). Given that WDN operational tolerances accept 89--91\% empirical coverage, and marginal CP achieves 89.4--90.2\% (Table~\ref{tab:ablation}), the computational trade-off favors marginal CP for real-time DT deployment. Ablation shows raw LSTM variance under-covers by 6.8\%, confirming CP necessity over heuristic uncertainty estimates.

\subsection{Sensitivity Analysis}

We varied the conformal level $\alpha$ and the lookback window $L$ (Table~\ref{tab:sensitivity}).
Lower $\alpha$ increases coverage (e.g., 93.4\% at $\alpha=0.05$) while slightly widening the intervals;
$L=24$ provides the best trade-off (RMSE $0.037~\mathrm{L/s}$ and about 90\% coverage).

\begin{table}[htb]
\scriptsize
\centering
\caption{Sensitivity of performance metrics to $\alpha$ and $L$.}
\label{tab:sensitivity}
\begin{tabular}{cccc}
\hline
$\alpha$ & $L$ & RMSE (L/s) & Coverage (\%) \\ \hline
0.05 & 24 & 0.039 & 93.4 \\
0.10 & 24 & 0.037 & 90.1 \\
0.20 & 24 & 0.040 & 87.6 \\
0.10 & 12 & 0.041 & 89.7 \\
0.10 & 48 & 0.038 & 90.3 \\ \hline
\end{tabular}
\end{table}

\subsection{Limitations and Future Work}

This study uses simulated data; real deployment may face sensor drift, delay, 
and network variation. Marginal conformal calibration lacks strict conditional 
coverage for time series, so future work will test sequential methods 
(SPCI~\cite{xu2023r}, KOWCPI~\cite{lee2025}) for stronger validity. The greedy policy may not be optimal; next steps include RL-based sampling, physics-informed models, comparison with GNN-based spatial uncertainty models, and pilot tests on real WDNs. 

\section{Conclusion}\label{sec:conclusion}

This paper proposed an adaptive sampling framework for DTs of WDNs using uncertainty from Conformal Prediction. By combining LSTM forecasting with dynamic sensor allocation, the method achieved about 33\% lower prediction error and maintained around 90\% empirical coverage on three benchmark networks. The adaptive strategy required only 5\% extra computation time compared with the hydraulic simulation, allowing real-time operation. Overall, the framework offers an efficient way to balance accuracy and sensing cost. Future work will address sensor noise, larger systems, and integration with control functions.

\section*{Acknowledgments}
This work is funded by the European Union (NextGenerationEU) through INCIBE under project C107/23, “Artificial Intelligence Applied to Cybersecurity in Critical Water and Sanitation Infrastructures.”

\section*{Data Availability} \label{sec:data_availability} The source code and all replication files are available at \url{https://github.com/Homaei/Conformal}.

\section*{Conflict of interest}
The authors declare that there is no conflict of interest in this paper.

\bibliographystyle{elsarticle-num}

\end{document}